**Chapter 1**

# Generic Ontology Design Patterns: Roles and Change Over Time

**Bernd Krieg-Brückner**, CRC EASE, Universität Bremen, and DFKI, Germany

**Till Mossakowski**, Otto-von-Guericke-Universität Magdeburg, Germany

**Mihai Codescu**, CRC EASE, Universität Bremen, Germany[1]

## 1.1. Introduction

In ontology engineering, we may distinguish (at least) three kinds of stakeholders: *ontology experts, domain experts* and *end-users*. For each, the kind and level of expertise is quite different: End-users should not be required to have ontology expertise and may have little domain knowledge; domain experts are likey to have little ontology expertise; ontology experts usually have little specialised domain expertise.

Since many ontology development projects are small and, frequently, not well funded, many of them do not involve a dedicated ontology expert at all or get only very limited input from an ontology expert during the design phase of the ontology. Thus, the majority of the development of an ontology is often entrusted to domain experts. This may lead to poor design choices and avoidable errors. Furthermore, because of their lack of experience, domain experts may not be able to identify opportunities to reuse existing best practices and ontologies.

Ontology Design Patterns (ODPs) have been proposed for some time as a methodology for ontology development, see the early work by [1, 2], the compilation in [3, 4, 5], and the review of the state of the art in [6].

In theory, ODPs provide a solution for the lack of ontology experts: ODPs enable domain ontologists to reuse existing best practices and design decisions, and, thus, benefit from the experience of ontology experts, who developed the ODPs. However, in practice the adaptation of ODPs as tools for ontology engineers has been slow. In our opinion this is caused by the fact that currently the utilisation of ODPs is cumbersome for ontology developers.

[1]The research reported in this paper was partially supported by the German Research Foundation (DFG), as part of the Collaborative Research Center (Sonderforschungsbereich) 1320 "EASE - Everyday Activity Science and Engineering" (http://www.ease-crc.org/),

In this chapter we propose *Generic Ontology Design Patterns*, GODPs, as a methodology for representing and instantiating ODPs in a way that is adaptable, and allows domain experts (and other users) to safely use ODPs without cluttering their ontologies. The main ideas behind are the following: ODPs are expressed in a dedicated formal, parameterised pattern language that allows (a) the definition of ODPs, (b) to specify instantiations of ODPs, (c) to extend, modify, and combine ODPs to larger ODPs, and (d) to describe the relationships between ODPs.

GODPs aid the three kinds of stakeholders mentioned above in the engineering of ontologies at various levels, cf. [7]: ontology experts develop a repository of foundational GODPs, to be used by well-trained domain experts (in cooperation with ontology experts) to configure a tool-box of domain-oriented GODPs dedicated to particular development tasks; end-users use this tool-box with an appropriate user interface for safely populating a data base, due to corresponding data integrity constraints.

GODPs are defined in Generic DOL, an extension of the *Distributed Ontology, Model and Specification Language*, DOL, and implemented in the *Heterogeneous Tool Set*, Hets. Generic DOL has special features such as *parameterized names* and *list parameters with recursion*, considerably facilitating the developer's work.

GODPs are patterns: they contain typed variables. The definition of a GODP involves the specification of parameters that need to be provided for the instantiation of a GODP. Parameters are *ontologies*; the case of symbols as parameters is covered by ontologies without axioms and only one symbol declaration. In general, ontology parameters enable the expression of *powerful semantic constraints* using corresponding axioms; such requirements act like preconditions for instantiations, guaranteeing more consistency and safety. For each argument for a parameter, a verification condition is generated. If expressed in description logic, DL (as in this paper), it may be discarded automatically by deduction using a DL reasoner; in a heterogeneous setting, DOL and Hets allow more expressive logics. We show in [7], how data consistency scales up with large amounts of data. In this context, formal parameters act as *data integrity constraints*.

We will introduce GODPs by representing two ODPs from the literature, and illustrate the instantiation of GODPs with examples. As we will show, GODPs enable the nested use of ODPs, reducing code duplication. Furthermore, GODP developers may explicitly state logical properties of GODPs, represent competency questions, and define extensions. During reuse, instantiations provide extra handles for improving consistency, checking arguments against requirements stated in the parameters.

GODPs share many objectives with *Parameterized OTTR Templates* with macro expansion [8], see Ch. 2 of this volume. In OTTR, a higher-order concept may be used for a list parameter: a template may be applied to all elements of a list, and lists may be combined; however, no general recursion is available as in Generic DOL. There are no ontology parameters or parameterised names in OTTR (cf. Sect. 1.3.1).

The authors of OTTR advocate templates everywhere, even small templates encapsulating just a single axiom. This enables their impressive approach to fold multiple occurrences of larger fragments into the instantiation of a template that is thus reused, see [9]. While this is useful for discovering new templates, we propose to only introduce a GODP, if it is very likely to be reused and encapsulates a larger fragment or an important design decision that might be revoked.

*Summary.* After a short review of DOL, Hets, and Generic DOL in Sect. 1.2, we introduce some simple GODPs. The features of Generic DOL are introduced informally with the examples, as we go along.

Sect. 1.3 shows, how two ODPs from the literature, the *Role* and the *Change Over Time* patterns (see Ch. 3 in this volume) may be compared, sharing very similar patterns; moreover, the latter provides additional auxiliary patterns, e.g. for *overloading invariant properties*, that can be conveniently reused for the former.

In Sect. 1.4, these patterns are then used in a larger example, *Vehicles and Drivers*, to demonstrate their repeated use. We develop more patterns for *qualitative value sets*, *grades*, *graded relations* and their combination, etc., which, while useful in other applications (cf. [10]), allow us to model the intricate interrelation between PotentialDrivers properly licensed by a required DrivingLicence, and a RoadVehicle they want to (legally) drive. The crucial question is, how to express the requirement ensuring that a Driver only drives a particular RoadVehicle, if s/he is licencedFor it.

In Sect. 1.5 we are then able to show, as a "best practice", how this consistency requirement can be stated as a *precondition* on a GODP such that its instantiation generates a corresponding proof obligation to ensure adherence to this *data integrity constraint* (stated in OWL-DL, thus deduction terminates); *requirement checking* already happens "statically" *during the Generic DOL expansion and verification process* in Hets.

We conclude in Sect. 1.6 with some objectives for future work.

**1.2. Generic Ontology Design Patterns in** Generic DOL

*1.2.1. DOL and Hets, Generic DOL*

The *Distributed Ontology, Model and Specification Language*, DOL, an OMG standard [11, 12, 13], allows the structured definition of ontologies, using import, union, renaming, module extraction, and many more. Thus, DOL is not yet another ontology language, but a meta-language, which allows the definition and manipulation of ontologies on the level of structuring. DOL can be used on top of a variety of languages, in particular OWL 2. It is supported by the *Heterogeneous Tool Set*, Hets [14], providing parsers for DOL specifications, an implementation of DOL semantics, and an interface to theorem provers.

*Unions and Extensions in DOL.* The building blocks of DOL are (basic) ontologies written as-is in existing ontology languages like OWL, or Common Logic, inheriting their semantics. DOL also provides a construct for *uniting* ontologies, written $O_1$ **and** $O_2$, and one for *extending* an existing ontology $O_1$ by new declarations and axioms, written $O_1$ **then** $O_2$ (in this case, $O_2$ may be an ontology fragment that is only well-formed in the context of $O_1$). When all basic ontologies are written in OWL, this has the same expressivity as OWL imports.

Note that when using DOL unions or extensions in an unrestricted way, it is possible that the resulting ontology, when flattened, is no longer in OWL, even if the building blocks are. (The same can happen with OWL imports already, so this is not particular to DOL.) The reason is that certain combinations of axioms are forbidden in OWL. Unrestricted union or extension can therefore lead to ontologies formulated in *OWL without*

```
pattern TotalRELATION_ScopedRange [ObjectProperty: p; Class: D; Class: R] =
  Class: D SubClassOf: p some R and p only R
```

**Figure 1.** TotalRELATION_ScopedRange

```
pattern TEMPORAL_Extent [Class: C]
=     TotalRELATION_ScopedRange [hasTemporalExtent; C; TemporalExtent]
then Class: TemporalExtent DisjointWith: C

ontology TEMPORAL_Extent_Vehicle_log =
    TEMPORAL_Extent [Vehicle]

%% ... expansion of the above instantiation to OWL:
ObjectProperty: hasTemporalExtent
Class: TemporalExtent DisjointWith: Vehicle
Class: Vehicle  SubClassOf: hasTemporalExtent only TemporalExtent
                       and hasTemporalExtent some TemporalExtent
```

**Figure 2.** TEMPORAL_Extent, instantiation, and expansion to OWL

*restrictions*, see for [15] discussion and the idea of using hybrid OWL/FOL reasoning for such ontologies.

The language Generic DOL was proposed in [16] as an extension of DOL with parameterized ontologies (i.e. generic ontologies, or patterns, for short), following generic specifications in CASL [17, 18], and used for writing GODPs.

[19] introduces important extensions such as lists, recursion, and local patterns, and also provides an informal overview of the Generic DOL semantics. We refrain from repeating this here and instead introduce the language with the help of examples.

*1.2.2. Generic Ontology Design Patterns*

*Scoped Relation.* Let us consider some simple GODPs first. In Fig. 1, the keyword **pattern** introduces a declaration of the pattern named TotalRELATION_ScopedRange with three parameters, written in square brackets, separated by "**;**": (the name for) an object property p, and classes D and R capturing its domain and range. These parameters may be instantiated later on with different names in arguments.

The body of a pattern (after the "**=**") is a Generic DOL specification making use of the symbols declared in the parameters (see [19] in general for further semantics detail). In the case of TotalRELATION_ScopedRange, the body is just an unstructured OWL ontology in OWL Manchester Syntax (other ontology languages are available, e.g. FOL-based syntaxes). The body declares a scoped range in the sense of [20]. More precisely, it ensures that the range of p scoped with domain D is R. This leaves freedom for letting p have other ranges when scoped with different domains. Moreover, the body also ensures that p is left-total on D (in case that p is a function, this means it has to be a total function). The body of TotalRELATION_ScopedRange encapsulates totality and scoping; a user not proficient in all semantic details may rely on the specialists to have provided a carefully developed and proven solution that may be reused without having to think about it further.

*Temporal Extent.* The pattern TEMPORAL_Extent in Fig. 2 takes as parameter a class C and uses it as one of the arguments of an instantiation of the pattern TotalRELATION_ScopedRange[2]. The instantiation of a pattern is written by giving, after the name of the pattern, the list of arguments, between square brackets and separated by semicolons. In the body of TEMPORAL_Extent, the instantiation of TotalRELATION_ScopedRange is then further extended with a disjointness axiom between TemporalExtent and the parameter C.

The ontology TEMPORAL_Extent_Vehicle_log shows an instantiation of TEMPORAL_Extent. The argument is again an implicitly defined class Vehicle. The semantics of the instantiation is that all occurences of C in the body of TEMPORAL_Extent are replaced with the argument Vehicle. Since the body of TEMPORAL_Extent contains the instantiation of the pattern TotalRELATION_ScopedRange, the latter is replaced with the body of TotalRELATION_ScopedRange where the parameters p, D and R are replaced with the arguments hasTemporalExtent, Vehicle and TemporalExtent, respectively. The resulting OWL ontology is shown in Fig. 2. [3]

## 1.3. The Role and the Change Over Time Patterns

In this section we describe the use of GODPs for the structuring of the *Role* and *Change Over Time patterns*. We start with an auxiliary pattern first.

*Functionality* There are various ways to express functionality:

*Cautious Scoping in Scoped_FUNCTION_Inverse ( Fig. 3):* the axioms introduce a scoped domain and range for f in the sense of [20]. Moreover, f is required to be a total function on D. f is neither required to be total nor a function elsewhere, and f may be overloaded on other domains and ranges (especially if these are disjoint to D and R, resp.; cf. Sect. 1.3.2). The pattern is developed in a modular way: first a total scoped relation is specified, then this is extended to a scoped function and finally to a scoped function with inverse.

*Overloading in Overall_FUNCTION_Inverse (Fig. 3) :* the axiom Characteristics: Functional on f states that f is a partial function across all domains. Since no explicit domain and range restrictions are given, f may be overloaded.

*Comment only:* restrictions in DL to ensure termination for reasoners (e.g. when using f with a subproperty chain axiom),[4] prevent the use of axioms for functionality such as those in the above variants; however, the intent to express functionality is encapsulated in the GODP and documented with the name (cf. [7]). The pattern is omitted here.

---

[2]Details about conventions for instantiations can be found in [19].

[3]In DOL (and OWL), the *"Same Name–Same Thing" principle* is used: the definition of an entity may be repeated without introducing multiple occurrences of that entity. For Generic DOL, this means that if the body of a GODP declares an entity, the union of multiple instantiations of that GODP will contain only one occurrence of that entity. If this was not the intention, the entity should rather become a parameter of the GODP, such that each instantiation can assign it a different name.

[4]https://www.w3.org/TR/owl2-syntax/#Global_Restrictions_on_Axioms_in_OWL_2_D

```
pattern TotalRELATION_Scoped [ObjectProperty: p; Class: D; Class: R] =
    TotalRELATION_ScopedRange[p; D; R]
then Class: D EquivalentTo: p some R

pattern Scoped_FUNCTION [ObjectProperty: f; Class: D; Class: R] =
    TotalRELATION_Scoped [ f; D; R]
then Class: D SubClassOf: f max 1 R

pattern Scoped_FUNCTION_Inverse
 [ObjectProperty: f; Class: D; Class: R; ObjectProperty: finv] =
    Scoped_FUNCTION [f; D; R]
then ObjectProperty: finv InverseOf: f

pattern Overall_FUNCTION_Inverse
 [ObjectProperty: f; ObjectProperty: finv] =
    ObjectProperty: f    Characteristics: Functional
    ObjectProperty: finv InverseOf: f

pattern CLOSED_Scope [Class: D; Class: R; ObjectProperty: p] =
    Class: D EquivalentTo: (p some R) and (p only R)
```

**Figure 3.** Variants of scoped relations, functions and inverse; CLOSED_Scope

*Heterogeneous Approach:* functionality is expressed in another logic in DOL; then the pattern will not expand to an OWL-DL ontology. Hybrid OWL/FOL reasoning can be used, see the discussion of OWL without restrictions in Sect. 1.2.1. The pattern is too complex for our purposes and we omit it here.

Note that the structuring by GODPs comes in handy, since a change or choice of a variant can be localised to the body of one GODP, and, after replay of instantiations, applies to all its instances with immediate reuse.

*Closed Scope.* The pattern CLOSED_Scope is similar to TotalRELATION_Scoped-Range (Fig. 1). It expresses that (1) p is total on D with range R (D is a subclass of p some R), (2) the range of p scoped with domain D is R (D is a subclass of p only R) [20], and (3) D is maximal with respect to (1) and (2). (This maximality is a bit weaker than a scoped domain declaration; the latter would require that p some R is a subclass of D.) Functionality of p is not required. We will make use of this pattern later in the body of CHANGE_PD.

*Refinement.* The similarity between CLOSED_Scope and TotalRELATION_Scoped-Range can be captured formally using DOL's notion of *refinement* between ontologies and patterns. Let us assume that we have two ontologies O1 and O2 with the same set of entities (signature). Then O2 is a refinement of O1 if the axioms of O1 can be logically entailed from those of O2. In DOL this is written **refinement** R = O1 **refined to** O2. This postulates a relation between O1 and O2 that has to be verified by checking that the expected logical entailment holds. If the entities of O1 and O2 are not the same, then one must also specify as a part of the refinement, how the entities of O1 are mapped to entities of O2. As an example, if we instantiate the patterns TotalRELATION_Scoped and CLOSED_Scope with the same arguments, we can show that the instantiation of Total-

```
refinement R = CLOSED_Scope[ Class: D; Class: R; ObjectProperty: p ]
    refined to TotalRELATION_Scoped[ObjectProperty: p; Class: D; Class: R]
```

Figure 4. Refinement of patterns

```
pattern ROLE_Explicit [ Class: Rle;           %% id of role
  Class: Performer;                           %% Performer performs Rle
  ObjectProperty: performedBy;   ObjectProperty: performs;
? Class: Provider;                            %% Provider provides Rle
? ObjectProperty: providedBy;  ? ObjectProperty: provides]
=   Scoped_FUNCTION_Inverse[performedBy; Rle; Performer; performs]
 and Scoped_FUNCTION_Inverse[providedBy;  Rle; Provider;  provides]
 and TEMPORAL_Extent[Rle]

ontology Role_PotentialDriver_log =
ROLE_Explicit[PotentialDriver; Person; isPotentialDriverRoleOf;licencedAs;
                              DrivingLicence; hasLicence;   isLicenceOf]

pattern ROLE_Compact[ Class: Rle; Class: Performer; ? Class: Provider ] =
  ROLE_Explicit[Rle; Performer; performedBy[Performer]; performs[Rle];
                    Provider;  providedBy[Provider];   provides[Rle]]
```

Figure 5. ROLE_Explicit, Instantiation, and ROLE_Compact

RELATION_Scoped is a refinement of the instantiation of CLOSED_Scope. Since the refinement can be established for any choice of arguments, we can extend the concept to refinement between patterns and say that TotalRELATION_Scoped is a refinement of CLOSED_Scope.

Together with the obvious refinements through extensions in Fig. 3, we get the following chain of refinements:

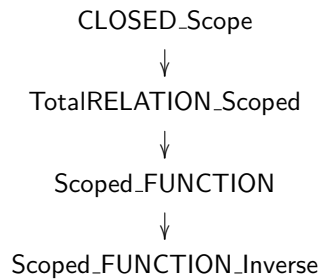

### 1.3.1. The Role Pattern

The Role ODP has received considerable attention in the literature, cf. e.g. [21]. How it can be turned into a structured GODP is described in [22] in some detail, comparing the ODP approach of subsumption (inheritance) to the GODP approach of parameterisation.

In [21], the object properties involved in the role pattern are axiomatised with an explicit domain, while the range (implicitly) is Thing. In [22], we generalise this to an arbitrary domain and range; the original pattern can be obtained by instantiating the range with Thing. In this chapter, we choose a different generalisation, namely we use *scoped* domains and ranges; again, the original pattern can be obtained by instantiating the scoped range with Thing.

ROLE_Explicit in Fig. 5 makes use of Scoped_FUNCTION_INVERSE twice, achieving a nice structuring and removing code duplication that has been present in the original ODP. Moreover, the use of a named pattern Scoped_FUNCTION_INVERSE makes the purpose of the axiomatisation more explicit. The two instantiations are combined with **and**, which is the union of ontologies.

A role Rle is performedBy a Performer, providedBy a Provider (where performs and provides are the corresponding inverses).

The variant ROLE_Explicit with 7 parameters allows maximal flexibility when giving names for arguments in special applications (cf. [7] and the examples below). Consider the instantiation Role_PotentialDriver_log (Fig. 5): special names such as licensedAs for performs are used, which convey better semantic connotations.

*Optional Parameters.* The parameters Provider, providedBy, and provides are *optional parameters*, indicated by the "**?**". If an empty argument is given, all axioms in the body referring to such optional parameters are rendered empty, and all instantiations in the body taking optional parameters become the empty ontology. The last axioms tie Rle to Performer and to Provider; the latter clause is empty if no argument for Provider is given.

*Parameterised Names.* In the instantiation of ROLE_Explicit in the body of ROLE_Compact, the argument perfomedBy[Perfomer] has a *parameterised name*. Parameterised names are a special feature of Generic DOL. The name perfomedBy[Perfomer] depends on the parameter Perfomer: different instantiations of ROLE_Compact with different arguments for them will give rise to different names for this class (analogously for performs[Rle], providedBy[Provider], and provides[Rle]). At the end of the expansion process of instantiations in Generic DOL, parameterized names are converted to proper OWL names by *stratification*: opening square brackets "[" and commas "," are turned into underscores "_", closing square brackets "]" are deleted.[5]

Consider the instantiation ROLE_Compact[MotherRole; Person; Childbirth]: since Rle is given the argument MotherRole, performs[Rle] expands to performs[MotherRole], and after stratification to performs_MotherRole. This seemingly innocuous feature considerably reduces parameter lists and simplifies instantiations and re-use, since every instantiation automatically generates a new set of (stratified) names.

ROLE_Compact (Fig. 5) is a compact specialisation variant of ROLE_Explicit. Apart from parameterised names, ROLE_Compact represents a faithful encapsulation of the original Role ODP (cf. [21]), structured with local instantiations of other GODPs; see the detailed description of its derivation in [22], where it corresponds to RoleGODPDecomposed. The parameterised names generate a separate name space for each instantiation; this is often preferable, renaming is avoided. If overloading is desired, e.g. for relation provides, then the more explicit pattern ROLE_Explicit should be used.

---

[5]Stratification makes nesting of square brackets in parameterised names (e.g. via nested instantiations) associative: name[A, B[C]], name[A[B], C] and name[A, B, C] will be equivalent after stratification.

```
pattern CHANGE_PD [ Class: C ] given Foundation =
    Overall_FUNCTION_Inverse[manifestationOf; hasManifestation]
and CLOSED_Scope[C;       PD[C];  manifestationOf]
and CLOSED_Scope[PD[C];  C;       hasManifestation]
and TEMPORAL_Extent[C]
then Class: C       SubClassOf: Manifestation
     Class: PD[C]  SubClassOf: TimeVaryingEntity

ontology Change_PD_Vehicle_log = CHANGE_PD[Vehicle]
  and OVERLOAD_Domain[PD[Vehicle]; hasManifestation; [VIN]; [hasVIN]]
  and SAME_Origin[Vehicle; PD[Vehicle]; manifestationOf]

%% ... expansion to OWL with stratification (without TemporalExtent):
Class: Manifestation    Class: TimeVaryingEntity
Class: VIN
Class: PD_Vehicle
       SubClassOf: hasVIN some VIN and hasVIN only VIN
       SubClassOf: TimeVaryingEntity
       EquivalentTo: hasManifestation some Vehicle
                     and hasManifestation only Vehicle
Class: Vehicle
       SubClassOf: Manifestation
       EquivalentTo: manifestationOf some PD_Vehicle
                     and manifestationOf only PD_Vehicle
ObjectProperty: hasManifestation
                InverseOf: manifestationOf
ObjectProperty: hasVIN
                SubPropertyChain: hasManifestation o hasVIN
                SubPropertyChain: inverse hasManifestation o hasVIN
ObjectProperty: manifestationOf
                Characteristics: Functional
ObjectProperty: same_PD_Vehicle
                Domain: Vehicle
                Range: Vehicle
                SubPropertyChain: manifestationOf o inverse manifestationOf
```

**Figure 6.** CHANGE_PD, instantiation and expansion to OWL (see Fig. 2 for TEMPORAL_Extent)

### 1.3.2. The Change Over Time Pattern

We now turn to another example of an ODP from the literature that we investigated. We started from the *Change Over Time* pattern in Ch. 3 in this volume and discovered a surprising amount of shared axiomswith the *Role* pattern; this lead to an attempt to compare and share GODPs for a maximal overlap of the two patterns, see below.

In Ch. 3 in this volume, an ODP for introducing change over time into an atemporal ontology is introduced. The major difference to the role pattern is that the former is more or less an ontology, while the change over time pattern is a "recipe" for adding time to an atemporal ontology. Such an addition requires non-trivial transformations, because some properties may become time-dependent, which leads to changes in both the signature and the axiomatisation of the atemporal ontology.

```
pattern CHANGE_Mf_Role [ Class: X ] =
  ROLE_Explicit[Mf[X]; X; manifestationOf; hasManifestation; ; ; ]
```

**Figure 7.** Change_Mf_Role

The approach of Ch. 3 is that objects changing in time (or *time varying entities*) are modeled as *perdurants* (in the terminology of the upper ontology DOLCE), i.e. they are identified with their life-spans. Objects at a specific time are considered as temporal parts of such a process; these temporal parts are called *manifestation*s.

CHANGE_PD in Fig. 6 is a GODP that should faithfully represent the "recipe" pattern for Change Over Time in Ch. 3 (Fig. 3.1); it formalises steps 1, 2, 4 and 5 of the recipe. CHANGE_PD adds a perdurant PD[C] to an existing class C, the latter corresponding to the manifestations. Applying this recipe to a given class C can now be easily obtained by instantiating CHANGE_PD with some argument class for C.

*Imports.* The last part of the body of CHANGE_PD states that the parameter C is a sub-class of a class Manifestation and that the class PD[C] introduced via the second instantiation of CLOSED_Scope is a sub-class of a class TimeVaryingEntity. Both these super-classes have been declared in an ontology called Foundation, that is *imported* by the pattern CHANGE_PD, using the **given** clause. An ontology imported by a pattern provides a context for it, in the sense that its entities are visible both in the parameters and in the body and thus may be involved both in the constraints and in the axioms generated by the pattern.

*Comparison to the Role Pattern.* When we first compared the CHANGE_PD pattern to the Role pattern, we found a surprising amount of shared axioms (cf. the refinement of CLOSED_Scope to TotalRELATION_Scoped above). Morever, our analysis revealed an accidental omission in the original version [23] of mentioning functionality for manifestationOf, which has since been remedied in Ch. 3 in this volume. The All_FUNCTION_INVERSE variant (cf. Sect. 1.3) is used to express functionality, since manifestationOf is intended to be functional across all domains. Scoped_FUNCTION_INVERSE in ROLE_Explicit is a more flexible solution allowing different names for performedBy that are not overloaded.

As shown by the refinement above, Scoped_FUNCTION_INVERSE (which is a part of ROLE_Explicit) is stronger than CLOSED_Scope (which is part of CHANGE_PD). However, since CHANGE_PD involves the pattern CLOSED_Scope twice (in two directions, cf. Fig. 3), in the end, this axiomatisation ties C and PD[C] much stronger together than ROLE_Explicit would do. More specifically, suppose A, B and PD[A], PD[B] denote classes and their perdurants, resp. Then, if A and B are declared to be disjoint, PD[A] and PD[B] are implicitly disjoint as well; they must not have a common subclass (or indeed a common individual). At first glance, it seems attractive to enforce such a separation in the pattern CHANGE_PD itself rather than moving the responsibility to take care of disjointness outside the pattern, as ROLE_Explicit does. However, in the terminology of roles, this would prevent that a Person may have two separate roles that are then set as disjoint, e.g. that a Driver must not be a ResidentPatient at the same time. Thus we prefer the weaker formulation of ROLE_Explicit as the more flexible solution.

```
pattern OVERLOAD_Domain [Class: D;        %% new domain
 ObjectProperty: map Domain: D;           %% from new to old domain
 Class: R :: Rs;                          %% list of ranges
 {ObjectProperty: p Range: R} :: ps] %% list of properties
= %% overloads invariant properties from old domain to new domain D
  ObjectProperty: p SubPropertyChain: map o p
                    SubPropertyChain: inverse map o p
then TotalRELATION_ScopedRange[p; D; R]
 and OVERLOAD_Domain[D; map; Rs; ps]

pattern SAME_Origin [Class: D; Class: R;
              ObjectProperty: f Domain: D Range: R] =
  ObjectProperty: same[R] Domain: D Range: D
    SubPropertyChain: f o inverse f
```

**Figure 8.** OVERLOAD_Domain and SAME_Origin

*Manifestations as Roles.* We might want to introduce a pattern that is dual to CHANGE_PD in going the other way, introducing a manifestation Mf[X] for an existing class X, and leaving X untouched as the corresponding perdurant. This variant is useful in ontology development, when an existing class with invariant properties exists (its invariant properties to be carried over with overloading, see Sect. 1.3.3), and new variant properties are to be defined subsequently for the new manifestation. In fact, apart from the differences compared above, this is equivalent to introducing a new role, with X as its "performer", more precisely to the "upper half" of the ROLE_Explicit pattern (Fig. 5), see CHANGE_Mf_Role in Fig. 7, where no arguments are given for a "provider" and the associated properties (optional arguments).

In the examples in the sequel, we will use ROLE_Explicit (Sect. 1.4, Fig. 9 ff.).

*1.3.3. Further Auxiliary Patterns for the Change Over Time and the Role Pattern*

The following patterns, inspired by Ch. 3 in this volume, are useful for both, the Change Over Time and the Role Pattern, to be used along with either.

*Overloading Properties.* In Ch. 3, properties already defined for C are separated into "variant properties" with the domain C, which are only meaningful and usable on C since they may change from one manifestation to another, and "invariant properties" that should be set on PD[C] and are invariant over all these manifestations. Since it is assumed there that we start from C, these properties need to be transferred to PD[C]. As a side-effect, it would be nice if they could be directly accessed from C as well, i.e. overloaded on both.

OVERLOAD_Domain (Fig. 8) provides a solution. For a new domain D, it takes a given relation p with range R, and a mapping map from D to the domain of p, and constructs a relation p from D to R, overloaded with the same name p, carrying over facts from the original p into induced facts from D to R. We can represent this as in the diagram below, where the dashed arrow is obtained from the solid arrows:

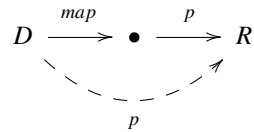

In contrast to Ch. 3 in this volume, where the general idea stems from, we also introduce the second subproperty chain in the inverse direction. It may happen, for example, that a fact for an invariant property can only be set late (e.g. the end date of some temporal extent, which can initially not be foreseen); then it can be set with either variant of p (i.e. for the old or the new domain) and also holds for the other, resp.

*Ontology Parameter for Typing Restriction.* Moreover, a special contribution of Generic DOL is used in OVERLOAD_Domain: semantic constraints on the parameters (for data integrity) are specified in the form of ontology parameters with axioms. Here, p is required to have the range R. We call such a simple constraint a "typing restriction" in analogy to the type analysis in programming languages. Instantiations have to conform to this constraint.

In general, parameters in Generic DOL are themselves ontologies. While a parameter may contain just the declaration of a symbol (whose kind in OWL is one of Class, Individual, ObjectProperty, or DataProperty), it may also be an arbitrarily *complex ontology* that contains axioms specifying specific abstract properties. An argument ontology must conform to such a parameter ontology, i.e. for simple parameters, without axioms, a symbol of the apropriate kind must be provided, and for parameters with axioms, the required properties must be satisfied by the argument. This amounts to checking that the argument is a refinement of the parameter, along an appropiate mapping of symbols of the parameter in the argument. Hets will take care of generating an appropriate proof obligation that will be checked by an automatic DL reasoner. Thus legality checking happens "statically" at DOL verification time (in analogy to static analysis for programming languages). This concept makes Generic DOL, and GODPs, extremely powerful to capture semantic preconditions for instantiations (cf. Sect. 1.5).

*List Parameters.* The above is generalised to lists in OVERLOAD_Domain: instead of a single parameter p, a list of relations is provided. The head of the list contains an object property p of range R, a typing restriction. For lists, an axiom stated in the head of the list is expected to hold for the elements of the tail of the list, ps, as well. The list constructor is written "::". Analogously, R is generalised to the list parameter Class: R :: Rs. The pattern is recursive, using as arguments the tails of these two lists; the implicit requirement is that the two lists must have the same length. The recursion ends when both lists are empty; then it is implicitly understood that the pattern results in an empty ontology.

*SAME_Origin.* The idea for SAME_Origin (Fig. 8) also stems from Ch. 3. Due to the instantiation in Change_PD_Vehicle_log (Fig. 6), we may inquire whether a particular manifestation Vehicle x is in fact the same[PD[Vehicle]] as a manifestation Vehicle y, i.e. whether they are manifestations of the same perdurant. Analogously for the Role pattern, we may inquire (Fig. 9), whether a particular PotentialDriver x is the same[Person] as a PotentialDriver y, e.g. if a DrivingLicence has been revoked and another re-issued later.

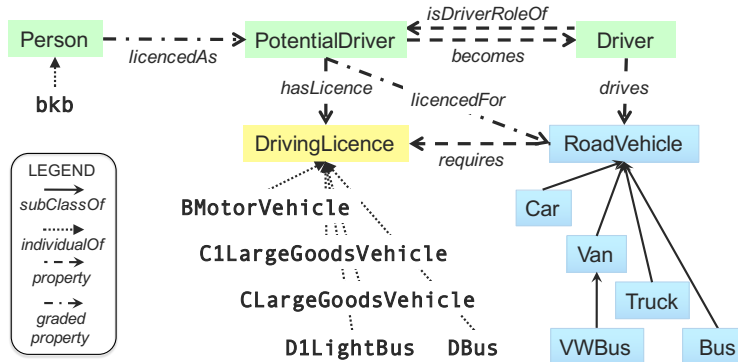

```
ontology Roles_Driver_log = Role_PotentialDriver_log
and ROLE_Explicit[Driver; PotentialDriver; isDriverRoleOf; becomes;
                  RoadVehicle;        drives;            drivenBy]
and OVERLOAD_Domain[Driver;isDriverRoleOf; [DrivingLicence]; [hasLicence]]
and SAME_Origin[PotentialDriver; Person; isPotentialDriverRoleOf]
and OVERLOAD_Domain[PotentialDriver; hasLicence;
                    [TemporalExtent]; [hasTemporalExtent]   ]
```

**Figure 9.** Vehicles and Drivers: Roles, Driving Licence, Vehicle Registration, etc.

SAME_Origin uses a parameterised name same[R].[6] This way, names may be kept separate in different instantiations, avoiding confusion from accidental overloading.

We may wish to add the clause "and SAME_Origin[C; PD[C]; manifestationOf]" inside the the CHANGE_PD pattern, or analogously to ROLE_Explicit for the Performer (it does not seem to make sense for the Provider).

## 1.4. Vehicles and Drivers

In this section we present a detailed example with some non-trivial patterns to show the power of Generic DOL, preparing for an interesting data integrity constraint in Sect. 1.5.

*1.4.1. Driver Roles.*

Fig. 9 shows some instantiations of the ROLE_Explicit pattern (Fig. 5) to drivers. Note that the instantiation of the pattern is iterated: a Person is licencedAs (i.e. performs the role of) a PotentialDriver when having acquired a DrivingLicence; then such a PotentialDriver becomes (i.e. performs as) a Driver when it drives a RoadVehicle. These roles are triggered by (i.e. providedBy) the corresponding contexts (DrivingLicence, RoadVehicle), and properly contain each other "in time": a PotentialDriver may sequentially assume the role of a Driver many times while the PotentialDriver role lasts (i.e. the DrivingLicence is not revoked or expires); when temporarily suspended, it may be reinstated later.

---

[6]same[R] only makes sense, if we can assume that f is functional; however, we did not require this explicitly in the parameter due to the OWL-DL restrictions.

We may now reuse the pattern OVERLOAD_Domain, i.e. the overloading idea from Ch. 3 (see Sect. 1.3.3), for roles. The property has_Licence is first introduced for the role PotentialDriver and may be overloaded now for Driver for direct access:

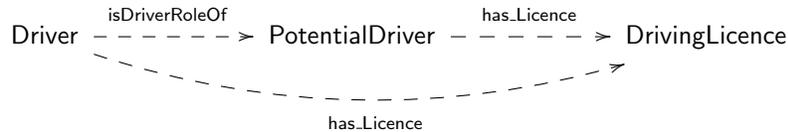

Similarly, other properties might be usefully overloaded (not shown), synchronously extending the singleton lists [DrivingLicence] and [hasLicence].

*Overloading Temporal Extent.* The issuing of the DrivingLicence provides the temporal context for the existence of a properly licenced PotentialDriver, cf. the OVERLOAD_Domain instantiation for TemporalExtent in Roles_Driver_log (Fig. 9): effectively, the TemporalExtent of DrivingLicence is inherited to PotentialDriver. In fact, since the Perfomer is most often related to a timed event, the inheritance of the TemporalExtent properties from Perfomer via overloading could bedirectly represented in the ROLE_Explicit pattern (Fig. 5). [7]

*Properly Licenced Driver.* The *crucial question* is, how to ensure that a Driver only drives a particular RoadVehicle, if s/he is licencedFor it, i.e. has an appropriate DrivingLicence for a RoadVehicle with certain constraints that the DrivingLicence requires.

The conventional way to achieve this would be to define subclasses of RoadVehicle, i.e. Car to correspond to BMotorVehicle, Truck to CLargeGoodsVehicle, and so on; cf. the apparent analogy in Fig. 9. However, it would be cumbersome to impose the structure of DrivingLicence onto the structure of RoadVehicle, in fact misleading in commonly used terminology, and thus dangerous. For example, a Van is not necessarily a small LargeGoodsVehicle (in the definition that the EU DrivingLicence `C1LargeGoods-Vehicle` requires), and a VWBus is definitely not a LightBus but a small kind of Van. It would be preferable to be completely free from potential licence constraints in defining the kinds of RoadVehicles, modelling DrivingLicence constraints separately. We will attempt to do so in the sequel, and prepare for it with more patterns in the GODP toolbox.

*Order Relations.* Only the axiom Characteristics: Transitive is defined in the body of Strict_ORDER (Fig. 10), more axioms for a strict order would be appropriate; unfortunately, restrictions of OWL-DL do not allow such an axiom, if r is ever to be used with a property chain or subproperty axiom (cf. Sect. 1.3). ORDER_Relations compactly defines order relations on X with parameterised names gt[X], ge[X], le[X] and lt[X], resp., thus distinguishing them from similar ones on other classes.

---

[7] The notion of time in the ROLE_Explicit or CHANGE_PD patterns is more sophisticated than just the sequential aspect for Episodes in [7]. Here, a TemporalExtent of a role (or manifestation) corresponds to a time interval; due to repeated instantiations, TemporalExtents may include each other, overlap etc., i.e. conform to some temporal interval algebra (still abstracting away from further details of real-time properties). TEMPORAL_Extent (cf. Fig. 2) suffices as a kind of stub for such concerns.

```
pattern Strict_ORDER [Class: X; ObjectProperty: r] =
 ObjectProperty: r Domain: X Range: X Characteristics: Transitive

pattern ORDER_Relations [Class: X] =
    Strict_ORDER[X; gt[X]] then
 ObjectProperty: ge[X] Domain: X Range: X Characteristics: Transitive
 ObjectProperty: gt[X] SubPropertyOf: ge[X]
 ObjectProperty: le[X] Domain: X Range: X Characteristics: Transitive
                       InverseOf: ge[X]
 ObjectProperty: lt[X] SubPropertyOf: le[X]
                       InverseOf: gt[X]
```

Figure 10. Strict_ORDER and ORDER_Relations

```
pattern VAL_Set [ Class: Val;         %% id of class of values
       ? ObjectProperty: greater;     %% id of order relation on Val
             Individual: v0 :: vs ]   %% ids of values
= %% all v in v0::vs become members of Val, ordered by greater
let  pattern OrderStep [Individual: i; Individual: j :: js] =
         Individual: j Types: Val Facts: greater i
      then OrderStep[j; js]
in   Individual: v0 Types: Val
then Strict_ORDER[Val; greater] and OrderStep[v0; vs]
then DifferentIndividuals: v0, vs
     Class: Val EquivalentTo: {v0, vs}

pattern GRADE [ Class: Ancestor; %% ancestor in taxonomy
                Class: Grade;    %% id of class of grades
             Individual: g :: gs ] %% ids of grades
=    VAL_Set[Grade; ; g :: gs]
then Class: Grade SubClassOf: Ancestor

pattern OrdGRADE [ Class: Ancestor; %% ancestor in taxonomy
                   Class: Grade;    %% id of class of grades
                Individual: g :: gs ] %% ids of grades
=    VAL_Set[Grade; gt[Grade]; g :: gs] and ORDER_Relations[Grade]
then Class: Grade SubClassOf: Ancestor
```

Figure 11. VAL_Set, GRADE and OrdGRADE.

*1.4.2. Qualitative Value Sets and Grades*

Qualitative values, corresponding to abstractions from quantitative data, occur quite often in practice. Cognitive science shows that they are related to the human need for doing away with irrelevant detail (precision in this case); they allow us to simplify abstract reasoning. In the context of grading, we introduce operations for combining qualitative values, allowing a kind of abstract calculation (cf. [24]).

Pattern VAL_Set (Fig. 11) defines sets of named individuals as values of Val (cf. [19]). As list parameters are now available in Generic DOL (cf. Sect. 1.3.3), this definition is considerably simpler and more flexible than the original formulation in [16].

```
ontology OrdGRADE_MaxSeats =
OrdGRADE[ VehicleAttribute ; MaxSeats ; [ le9Seats , gt9to17Seats , gt17Seats ] ]

%% expansion
Class: VehicleAttribute
Individual: gt17Seats      Types: MaxSeats
   Facts: gt_MaxSeats gt9to17Seats
Individual: gt9to17Seats Types: MaxSeats
   Facts: gt_MaxSeats le9Seats
Individual: le9Seats       Types: MaxSeats
DifferentIndividuals: le9Seats , gt9to17Seats , gt17Seats
Class: MaxSeats SubClassOf: VehicleAttribute
   EquivalentTo: {le9Seats , gt9to17Seats , gt17Seats}
ObjectProperty: ge_MaxSeats
   Domain: MaxSeats Range: MaxSeats Characteristics: Transitive
ObjectProperty: gt_MaxSeats
   SubPropertyOf: ge_MaxSeats
   Domain: MaxSeats Range: MaxSeats Characteristics: Transitive
ObjectProperty: le_MaxSeats
   Domain: MaxSeats Range: MaxSeats
   Characteristics: Transitive      InverseOf: ge_MaxSeats
ObjectProperty: lt_MaxSeats
   SubPropertyOf: le_MaxSeats      InverseOf: gt_MaxSeats
```

**Figure 12.** Instantiation OrdGRADE_MaxSeats and expansion.

greater is an optional parameter in ValSet; thus Val may be an ordered set if desired, but need not. The instantiation of Strict_ORDER defines greater as a strict order relation (see Sect. 1.4.1, Fig. 10); it is void, if this optional argument is missing. The last two axioms of the pattern make use of the variable vs for the tail of the list parameter in a position where an individual is expected. At instantiation, the variable is replaced with the tail of the argument, and thus the expected enumeration of individuals is obtained.

*Nested Patterns.* Another special contribution of Generic DOL is the possibility to define locally nested patterns. In Fig. 11, the pattern OrderStep is defined in the body of VAL_Set, inside a **let DECLS in SPEC** block, where **DECLS** is a list of declarations of patterns and **SPEC** is a Generic DOL specification. The scope of a local subpattern occuring in **DECLS** is thus restricted to the declarations following it in **DECLS** and to the specification **SPEC**. All names of the entities of the parameters of VAL_Set are locally visible in OrderStep; if VAL_Set were not local to OrderStep, we would have to add these parameters to VAL_Set as well, making its definition more complicated. This, together with list parameters, considerably simplifies definition and use of GODPs compared with the initial definitions in [16]. Note that OrderStep is again defined recursively.

*Grades.* A grade set is an (unordered) value set, see GRADE in Fig. 11. The parameter Ancestor denotes the place in the taxonomic hierarchy to position Grade.

For unordered grades see DrivingLicence in the diagram of Fig. 9. The various abstract values, corresponding to (a subset of) the kinds of European Driving Licences, come with different restrictions as we will see below (where an order will later be imposed since one license may imply others w.r.t. permitted vehicle use).

```
pattern GRADED_Rels [Class: S; Class: T;      %% domain and range
 ObjectProperty: p Domain: S Range: T;        %% property to grade
 Class: Ancestor;                              %% Grade's ancestor
 Class: Grade;                                 %% grading class
 Individual: g :: gs]                          %% grading values
= let pattern GradedRels[Individual: v :: vs] =
        ObjectProperty: p[v] Domain: S Range: T SubPropertyOf: p
      then GradedRels[vs]
in  GRADE[Ancestor; Grade; g :: gs] and GradedRels[g :: gs]
```

**Figure 13.** GRADED_Rels

Analogously, OrdGRADE imposes an order. In Fig. 11 we see an instantiation: MaxSeats is a set of qualitative values denoting intervals le9Seats, gt9to17Seats, gt17Seats (see [10] for the abstraction from quantitative intervals to abstract values).

*1.4.3. Qualitatively Graded Relations*

Grading with qualitative values occurs for human abilities (e.g. severely, moderately, slightly reduced), but makes analogous sense in robotics and AI (cf. [10]). A typical example is the significance of an ingredient in a recipe, where the graded relations could be hasIngredient_Insignificant, hasIngredient_Subordinate, hasIngredient_Essential and hasIngredient_Dominant.

A general approach for grading in ontologies is the introduction of a *sheaf of graded relations* for a given relation [25]. This is needed because OWL only supports binary relations; therefore the third argument of the relation, giving its qualitative value, is encoded in the name of a graded relation that is part of the sheaf. Consider GRADED_Rels in Fig. 13 (cf. [19, 7]): for a given relation p and set of grades Grade, it introduces a sheaf of graded relations p[$g_i$], one for each $g_i$ in g :: gS via the iteration in the local pattern GradedRels. The intended meaning is that

$$p(?s,?t,g) \equiv p\_g(?s,?t)$$

for a ternary relation p with grade value g as third argument. GRADED_Rels instantiates GRADE with the proper arguments, in particular the list of grades g :: gs; this avoids a potentially error-prone repetition of the list. If an instantiation of GRADE has already been made, it is identified by the "Same Name–Same Thing" principle.

In the first instantiation of GRADED_Rels in Fig. 16, the relation licencedAs is graded into a sheaf of relations (licencedAs[BMotorVehicle], ..., licencedAs[DBus]) according to the DrivingLicence (cf. Fig. 9).

LeGRADED_Rels in Fig. 14 introduces a relation p[le[$g_i$]] for each p[$g_i$] according to an ordered grade with a relation le[Grade]. This subsumption of all relations with a lesser grading than a particular $g_i$ is very useful; in the instantiation of LeGRADED_Rels in Fig. 16, the modelling as licencedFor[le[dl]], where dl is a DrivingLicence, ensures that a higher licence also includes a lesser licence: e.g. licencedFor[le[CLargeGoodsVehicle]] transitively subsumes licencedFor[BMotorVehicle]. The pattern makes use of the pattern AND_nRels that introduces a new object property as a subproperty of all elements of a

```
pattern AND_nRels [Class: S; Class: T;           %% domain and range
   ObjectProperty: r;                             %% id of result
  {ObjectProperty: p Domain: S Range: T} :: ps] %% list of relations
= %% a r b -> (a p1 b /\ ... /\ a pn b) for all pi in p::ps
      ObjectProperty: r Domain: S Range: T SubPropertyOf: p
then AND_nRels[S; T; r; ps]

pattern LeGRADED_Rels [Class: S; Class: T; %% domain and range
  ObjectProperty: p Domain: S Range: T;    %% property to grade
           Class: Ancestor;                %% Grade's ancestor
           Class: Grade;                   %% grading class
           Individual: g :: gs  ]          %% grades
= let
 pattern LEInitial[Individual: x] =
  AND_nRels[S; T; p[le[x]]; [p[x]] ]
 pattern LEStep [Individual: x; Individual: y :: ys] =
  AND_nRels[S; T; p[le[y]]; [p[le[x]], p[y]] ] then LEStep[y; ys]
in    OrdGRADE[ Ancestor; Grade; g :: gs] and
      GRADED_Rels[S; T; p; Ancestor; Grade; g :: gs]
 and  LEInitial[g] and LEStep[g; gs]
```

**Figure 14.** LeGRADED_Rels

list of object properties, passed as a parameter. LeGRADED_Rels can easily be extended to an analogous LeGtGRADED_Rels that also introduces a p[gt[$g_i$]].

Tabular_AND_3 in Fig. 15 allows the definition of a combination of subproperty axioms in the form of a two-dimensional table that looks like a combination of grade values; in fact, such a combination is "lifted" to a combination of graded relations for each cell of the table. Instantiations of this pattern need to provide a two-dimensional table of individuals. Here, "{}" denotes the empty ontology; it represents "undefined" in the table and, in effect, expands the combination to an empty axiom. While the number of columns is flexible, for simplicity, we have fixed the number of rows to 3. However, the pattern could be generalized arbitrarily many rows.

The pattern Tabular_AND_3 can be best understood with the help of an example. Let us assume we have three object properties, called rx, ry and rz, all of domain S and range T. Further assume we have three types of grades, Gx, Gy and Gz, and three instances y0, y1 and y2 of Gy, a list [x0, x1] of instances of Gx, and three lists of instances of Gz, [a0, a1], [b0, b1] and [c0, c1]. Instantiated with these arguments, the pattern produces the graded relations in the table below, where the relations rz[X] are sub-properties of the relations occurring in their corresponding row and column:

|        | rx[x0] | rx[x1] |
|--------|--------|--------|
| ry[y0] | rz[a0] | rz[a1] |
| ry[y1] | rz[b0] | rz[b1] |
| ry[y2] | rz[c0] | rz[c1] |

*Drivers for Properly Licensed Vehicles.*   In Driver_log in Fig. 16 we are now ready to use the above toolbox of GODPs. The first three pattern instantiations introduce graded

```
pattern Tabular_AND_3
[ Class: S; Class: T; Class: Gradex; Class: Gradey; Class: Gradez;
  ObjectProperty: rx Domain: S Range: T;
  ObjectProperty: ry Domain: S Range: T;
  ObjectProperty: rz Domain: S Range: T;
                {Individual: x    Types: Gradex} :: xChain;
 {Individual: y0 Types: Gradey};
                {Individual: xy0 Types: Gradez} :: xy0Chain;
 {Individual: y1 Types: Gradey};
                {Individual: xy1 Types: Gradez} :: xy1Chain;
 {Individual: y2 Types: Gradey};
                {Individual: xy2 Types: Gradez} :: xy2Chain]
= %% (a rz[xiyj] b -> a rx[xi] b) and (a rz[xiyj] b -> a ry[yj] b)
    AND_nRels[S; T; rz[xy0]; [rx[x], ry[y0]]]
and AND_nRels[S; T; rz[xy1]; [rx[x], ry[y1]]]
and AND_nRels[S; T; rz[xy2]; [rx[x], ry[y2]]]
and Tabular_AND_3[S; T; Gradex; Gradey; Gradez; rx; ry; rz;
                  xChain; y0; xy0Chain; y1; xy1Chain; y2; xy2Chain]
```

**Figure 15.** Tabular_AND_3

relations for driver licence, for authorized mass and for the number of seats of a vehicle. The instantiation of LeGRADED_Rels ensures, as explained previously, that a higher licence also includes a lesser licence. We define the graded relation licencedFor a particular DrivingLicence as a combination of relation mightDrive with the graded MaxSeats restriction and the relation mightDrive with a graded MaxAuthorisedMass restriction. This is achived by the last instantiation of Tabular_AND_3. For example, we obtain that a PotentialDriver is suitably licencedFor[BMotorVehicle] a RoadVehicle with MaxSeats le9Seats and MaxAuthorisedMass le3500kg; similarly, a PotentialDriver is suitably licencedFor[D1LightBus] a RoadVehicle with MaxSeats gt9to17Seats and MaxAuthorisedMass gt3500to7500kg, but not gt7500kg (notice the "{}" in that position in the table). A BMotorVehicle licence would suffice for a VWBus with MaxSeats le9Seats and MaxAuthorisedMass le3500kg.

In future work, the Driver role could then be related to roles of RoadVehicles and other agents in traffic situations, in fact traffic Episodes as an adaptation of [7].

## 1.5. Data Patterns, Data Consistency

In Data_Driver_log (Fig. 17) we can see an instance of the data integrity constraint mentioned and informally stated in Sect. 1.4.1: The pattern DATA_Role expresses that an instance rle of a role class R is provided by an instance of the class Provider and an instance of the class Performer performs the role rle. This is specialised in the pattern DATA_Driver_Role, which allows a PotentialDriver pd to drive a RoadVehicle rv *only*, if pd is licencedForSomeDl rv, i.e. has an appropriate licence dl. The instantiation of DATA_Role ensures that bkb is licencedAs a PotentialDriver bkb_PotentialDriver with a BMotorVehicle licence (cf. Fig. 9). This is now stated as a *precondition* on pd, i.e. Individual: pd Facts: licencedForSomeDl rv. The ontology

```
ontology Driver_log = Roles_Driver_log
and GRADED_Rels[ Person; LicencedPerson; licencedAs;
     DrivingAttribute; DrivingLicence;
     [BMotorVehicle, C1LargeGoodsVehicle, CLargeGoodsVehicle,
      D1LightBus, DBus                                          ] ]
and GRADED_Rels[ PotentialDriver; RoadVehicle; mightDrive;
     VehicleAttribute; MaxAuthorisedMass;
     [le3500kg, gt3500to7500kg, gt7500kg] ]
and GRADED_Rels[ PotentialDriver; RoadVehicle; mightDrive;
     VehicleAttribute; MaxSeats;
     [le9Seats, gt9to17Seats, gt17Seats] ]
and LeGRADED_Rels[ PotentialDriver; RoadVehicle; licencedFor;
     DrivingAttribute; DrivingLicence;
     [BMotorVehicle, C1LargeGoodsVehicle, CLargeGoodsVehicle,
      D1LightBus, DBus                                          ] ]
and Tabular_AND_3[ PotentialDriver; RoadVehicle;
              MaxSeats; MaxAuthorisedMass; DrivingLicence;
              mightDrive; mightDrive; licencedFor;
           [    le3500kg,       gt3500to7500kg,        gt7500kg     ];
 le9Seats;    [BMotorVehicle,C1LargeGoodsVehicle,CLargeGoodsVehicle];
 gt9to17Seats;[    {},             D1LightBus,            {}        ];
 gt17Seats;   [    {},                DBus,              DBus       ]]
```

**Figure 16.** Driver_log: roles for drivers related to properly licensed vehicles (cf. Fig. 9)

Data_Driver_log shows an instantiation of the DATA_Driver_Role: bkb_PotentialDriver only becomes a legal bkb_BusDriver of bkbs_VWBus, because bkb_PotentialDriver is licencedFor[le[BMotorVehicle]] (since the previous instantiation ensures that it is licencedFor[BMotorVehicle]) and thus bkbs_VWBus, being a VWBus, complies with the restrictions of the licence BMotorVehicle as stated in the table (Fig. 16) (see above). Note that the ontology fragment presented as the last item in Fig. 17 consists only of the sentences from the body of DATA_Role instantiated via DATA_Driver_Role as specified in the last instantiation in the body of DATA_Driver_log. The full expansion contains more sentences, e.g. those coming from Driver_log. Moreover, the condition on the RoadVehicle bkbs_VWBus is stated via licencedFor[le[BMotorVehicle]], and that VWBus is a subclass of Van can be stated free from this restrictions (whereas if we had defined the licence BMotorVehicle to strictly correspond to Car, VWBus would have to be categorised as a Car).

*Checking Constraints at Verification Time.* Note that we have thus moved the check for legality into a *precondition of a GODP* such that its instantiation will generate a corresponding proof obligation to ensure that the data integrity constraints holds. Since this proof obligation is stated in OWL-DL, it can be taken care of automatically (with termination assured by OWL-DL). Thus legality checking happens at "DOL verification time" (in analogy to static analysis for programming languages); otherwise the checking for legality of driving would have to be relegated to an explicit event (i.e. a police officer would have to do the checking).

```
pattern DATA_Role [ Class: R; %% role class
Class: Performer;  ObjectProperty: performs    Domain: Performer Range: R;
Class: Provider;   ObjectProperty: providedBy  Domain: R Range: Provider;
Individual: perf; Individual: prov; Individual: rle ]
= Individual: prov Types: Provider
  Individual: rle  Types: R            Facts: providedBy prov
  Individual: perf Types: Performer    Facts: performs rle

pattern DATA_Driver_Role
[ ObjectProperty: licencedForSomeDl
                Domain: PotentialDriver Range: RoadVehicle;
  Individual: rv;
  Individual: pd  Facts: licencedForSomeDl rv;
  Individual: d                                   ] %% id of Driver
given Driver_log %% Driver, PotentialDriver, RoadVehicle, ...
= DATA_Role[Driver; PotentialDriver; becomes; RoadVehicle; isDriverOf;
          pd; rv; d]

ontology Data_Driver_log = Driver_log
and DATA_Role[PotentialDriver; Person; licencedAs; DrivingLicence;
             hasLicence; bkb; BMotorVehicle; bkb_PotentialDriver]
and DATA_Driver_Role[licencedFor[le[BMotorVehicle]];
                    bkbs_VWBus; bkb_PotentialDriver; bkb_BusDriver]

%% ... expansion of the last instantiation above:
Individual: bkb_PotentialDriver Types: PotentialDriver
                                Facts: becomes bkb_BusDriver
Individual: bkb_BusDriver       Types: Driver
                                Facts: isDriverOf bkbs_VWBus
```

**Figure 17.** DATA_Role, DATA_Driver_Role and instantiations in Data_Driver_log

### 1.6. Conclusion, Ongoing and Future Work

Starting with [25] and [16], with the extensions in [19], most GODPs shown here have been used in other non-trivial and large examples, e.g. in [10, 7], primarily in the robotics domain; some updates are presented here. These tried-out examples have demonstrated their usefulness; indeed, we relied on the successful completion of such examples supported by the inherent structuring of GODPs in the design, and the tool support of Hets during the development, providing essential static analysis and consistency checking.

The examples not only demonstrate the development methodology with GODPs and show the power of (list parameters and) ontology parameters to express non-trivial preconditions in OWL-DL, but also open a perspective for a new paradigm for modelling consistency of ontologies and integrity constraints for safe input of data by giving the opportunity of checking "statically" at DOL expansion and verification time.

*Change Management.* The log files, e.g. Data_Driver_log (Fig. 16), are ontologies that only contain instantiations of GODPs. They represent basic units for change management: whenever one of the GODPs, directly or indirectly used in the body, is changed, such a log file may easily be replayed to distribute the effect.

Moreover, such logs may themselves become objects ("first-order citizens") for treatment: with suitable meta-axioms (future work) we may reason about possible rearrangements, for example w.r.t. their order, if they are mutually independent.

*Acknowledgements.* We are very grateful to Fabian Neuhaus and Mihai Pomarlan for their suggestions and contributions. We would also like to thank Megan Katsumi and Mark Fox for their cooperation regarding the Change Over Time pattern.